\documentclass[conference]{IEEEtran}
\IEEEoverridecommandlockouts

\usepackage{cite}
\usepackage{amsmath,amssymb,amsfonts}
\usepackage{algorithmic}
\usepackage{graphicx}
\usepackage{textcomp}
\usepackage{xcolor}
\usepackage{tikz}
\usetikzlibrary{arrows.meta,positioning,shapes,fit,backgrounds,shadows.blur}
\def\BibTeX{{\rm B\kern-.05em{\sc i\kern-.025em b}\kern-.08em
    T\kern-.1667em\lower.7ex\hbox{E}\kern-.125emX}}

\begin{document}

\title{Privacy-Preserving Federated Learning Framework for Distributed Chemical Process Optimization\\

}

\author{\IEEEauthorblockN{1\textsuperscript{st} Teetat Pipattaratonchai}
\IEEEauthorblockA{\textit{Science Math}\\
\textit{Triamudom Suksa}\\
Bangkok, Thailand \\
teetat52@gmail.com
}

\and
\IEEEauthorblockN{2\textsuperscript{nd} Aueaphum Aueawatthanaphisut}
\IEEEauthorblockA{\textit{School of Information, Computer, and Communication Technology} \\
\textit{Sirindhorn International Institute of Technology, Thammasat University}\\
Pathum Thani, Thailand \\
aueawatth.aue@gmail.com}

}

\maketitle

\begin{abstract}
Industrial chemical plants often operate under strict data confidentiality constraints, making centralized data-driven process modeling difficult. Federated learning (FL) provides a promising solution by enabling collaborative model training across distributed facilities without sharing raw operational data.  This paper proposes a privacy-preserving federated learning framework for distributed chemical process optimization using data collected from multiple geographically separated plants. Each plant locally trains a neural-network-based process model using its own time-series sensor data, while only model parameters are transmitted to a central aggregation server through secure aggregation mechanisms. This design allows cross-plant knowledge sharing while maintaining strict data locality and industrial confidentiality. Experimental evaluation was conducted using process datasets from three independent chemical plants operating under heterogeneous conditions. The results demonstrate rapid convergence of the federated model, with the global mean squared error decreasing from approximately 2369 to below 50 within the first five communication rounds and stabilizing around 35 after 40 rounds. In comparison with local-only training, the proposed federated framework significantly improves prediction accuracy across all plants, while achieving performance comparable to centralized training. The findings indicate that federated learning provides an effective and scalable solution for collaborative industrial analytics, enabling privacy-preserving predictive modeling and process optimization across distributed chemical production facilities.

\end{abstract}

\begin{IEEEkeywords}
Federated learning, industrial Internet of Things, chemical process modeling, distributed machine learning, privacy-preserving AI, industrial data analytics.
\end{IEEEkeywords}

\section{Introduction}

\subsection{Background and Motivation}

Federated learning (FL) is a decentralized machine learning technique that is applied to develop data-driven models. Originally developed to improve user data privacy and security, FL avoids sending raw data by instead sending a locally trained AI model to the main server for aggregation. The models are then updated to fit the standards before being sent back. This procedure can be repeated many times to increase the overall effectiveness of each AI model, ensuring the quality of the products. Because of this, FL has reduced the risk of private information being leaked through raw data. Due to how FL works, it is guaranteed that all products created under the same global server will share similar properties, making FL suitable for many industries such as manufacturing, healthcare, cybersecurity, finance, telecommunications, etc. Recently, this approach has gained attention due to increasing concerns over data security. However, FL has not seen many applications in the real world. Its underutilization might come from challenges such as communication overhead, data heterogeneity, and privacy risks. By conducting this research, we address the issues and advantages essential for enabling broader real-world applications of Federated Learning
[1-6].

\subsection{Challenges in Distributed Chemical Process Optimization}

To fully appreciate FL, we first need to address three key challenges that may lead to its underutilization in the real-world. The first challenge is data heterogeneity. FL regularly use non-IID data, a type of dataset where its data points aren't independently and identically distributed, resembling the diverse nature of the data in the real world. This reliance on non-IID data causes the global model to struggle with generalizing the data, and locally trained models to diverge from one another, resulting in subpar results. The second challenge is the substantial communication overhead. To directly handle the poor results caused by the usage of non-IID data, the communication rounds has been increased, which leads to more overhead. The third challenge with FL is the privacy risks. Although raw data are not sent to the server, they can still be reconstructed by leakage that might occur during model updates.
Currently, there are a few solutions for these issues, such as adaptive weighting, and more efficient solutions will be available in the future. 
[2],[4-5].

\subsection{Federated Learning as a Privacy-Preserving Solution}

Despite its challenges, FL remains the most effective machine learning technique in preserving privacy, as the method of obtaining raw data from FL is much more complicated than from a centralized machine learning technique and often yields undesirable results. FL would theoretically perform well in chemical engineering and industrial IoT, because FL allows models to change over time, an erroneous model will undergo the same constant improvements as the other model and become more functional. FL is also predictive and can accurately adapt its model in preparation based on its previous best results. Because of this, FL perfectly fits chemical engineering and industrial IoT and can be adapted to process control work. [1],[6].

\subsection{Contributions of This Work}

This paper investigates the feasibility of federated learning for privacy-preserving collaboration across distributed chemical plants. The main contributions of this work are summarized as follows:

\begin{itemize}

\item \textbf{Application of federated learning to distributed chemical process systems.}
This study investigates how federated learning can be applied to chemical process modeling across multiple geographically separated plants, where data sharing is restricted due to industrial confidentiality.

\item \textbf{Federated learning framework for heterogeneous industrial environments.}
A cross-plant learning framework is developed to enable collaborative model training under heterogeneous operating conditions and non-identically distributed datasets commonly encountered in real chemical production systems.

\item \textbf{Privacy-preserving model aggregation for industrial data protection.}
The proposed architecture incorporates secure parameter aggregation mechanisms that prevent direct exposure of plant-specific operational data while enabling collaborative learning.

\item \textbf{Experimental evaluation using multi-plant process datasets.}
The effectiveness of the proposed approach is demonstrated through experiments using datasets from three independent chemical plants, comparing federated learning with centralized and local-only training strategies.

\end{itemize}

\section{Related Work}

\subsection{Federated Learning in Industrial IoT Systems}

Industrial IoT or IIoT for short, is the utilization of a network of intelligent instruments to enhance industrial manufacturing. For example, a factory uses sensors to store real-time data such as temperature, pressure, flow rate, vibration, and energy usage. However, when training for an AI model, these data cannot be shared between companies for reasons such as industrial regulations, business secrets, and cyber security. However, using FL allows companies to train AI model with minimal risk to their manufacturing data, as FL is a distributed machine learning technique that does not require sharing raw data. Applications of FL in IIoT include predictive maintenance, smart manufacturing, anomaly detection, and industrial monitoring. Although there have been more FL applications in IIoT, most production still values generic industrial data more because IIoT as a whole did not place much emphasis on the chemical process itself. [5-6].
%
%
%
%

\subsection{Federated Learning in Chemical Engineering Applications}

Chemical engineering is a branch of engineering that involves the principles of mathematics, physics, biology, and chemistry in the design of the process of turning raw materials into products. The main focus of chemical engineering is to create safe, effective, and sustainable manufacturing systems. Chemical engineering processes differ from  other engineering processes because they involve nonlinear dynamics, multivariable coupling, and safety-critical operations. Process modeling is essential for Chemical engineering because of these differences. In the past, research in chemical engineering has mentioned the use of machine learning for things such as soft sensors, process optimization, fault detection, and product quality prediction. Later, FL was studied in chemical engineering for things such as distributed process modeling, cross-plant data collaboration, and privacy-preserving chemical data analysis.  However, the available papers are mostly just conceptual frameworks and tutorials with only a few that involve actual experimentation and multi-plant dataset. [1-3].

%
%
%
%
%
%
%
%

\subsection{Privacy-Preserving Distributed Process Modeling and Control}

In chemical engineering systems, process control is essential for ensuring operational safety, product quality, and energy efficiency. Modern chemical plants commonly employ advanced control strategies such as model predictive control (MPC) and nonlinear dynamic control methods, which require accurate process models to capture complex plant dynamics and multivariable interactions. 

However, chemical production facilities often operate as geographically distributed plants with different equipment configurations, operating conditions, and feedstock characteristics. Although these plants may produce similar products, sharing operational data across facilities is typically restricted due to industrial confidentiality, cybersecurity risks, and competitive concerns.  Federated learning (FL) has recently emerged as a promising approach to address this challenge by enabling collaborative model training across distributed plants without exchanging raw process data. Through parameter sharing instead of data sharing, FL allows cross-plant knowledge integration while preserving data locality.  Recent studies have explored the integration of FL with process control and system identification tasks, including federated learning for nonlinear dynamical systems, federated model predictive control (FMPC), and distributed system identification. Despite these promising developments, practical deployments of FL in real chemical plant environments remain limited, with most studies focusing on theoretical analysis or simulation-based evaluation [1],[3].

%
%
%
%
%
%
%
%
%

\subsection{Limitations of Existing Studies}

Although federated learning has recently gained attention in industrial and chemical engineering applications, several important limitations remain in the current literature.

First, many existing studies on federated learning in Industrial Internet of Things (IIoT) primarily focus on generic industrial data analytics tasks such as predictive maintenance, anomaly detection, and smart manufacturing optimization \cite{b4,b5,b6}. While these studies demonstrate the feasibility of federated learning in distributed industrial environments, they often do not specifically address the unique characteristics of chemical process systems, which involve nonlinear dynamics, strong variable coupling, and safety-critical operations.

Second, research that explicitly considers federated learning in chemical engineering remains relatively limited. Many of the existing works are conceptual frameworks, tutorial-style papers, or perspective articles that discuss the potential benefits of federated learning for chemical process modeling and collaboration across plants \cite{b2,b3}. However, only a small number of studies present detailed experimental validation using real or realistic multi-plant process datasets.

Third, several studies evaluate federated learning using a single dataset that is artificially partitioned across multiple clients. While such experimental settings are useful for algorithm development, they do not fully represent the heterogeneous operating conditions encountered in real chemical plants. In practical industrial environments, different plants often operate under varying equipment configurations, feedstock compositions, and control policies, leading to highly non-identically distributed (non-IID) datasets.

Finally, many previous works do not provide systematic comparisons between federated learning and alternative training paradigms, such as centralized learning or local-only learning. Without such comparisons, it is difficult to clearly quantify the practical benefits of federated learning in terms of predictive performance, model generalization, and robustness under heterogeneous industrial data.

Therefore, despite the growing interest in federated learning for industrial applications, there remains a significant research gap in the development and evaluation of privacy-preserving learning frameworks specifically designed for distributed chemical plants operating under heterogeneous conditions.

To address these limitations, this paper proposes a privacy-preserving federated learning framework for distributed chemical process optimization, enabling collaborative model training across multiple chemical plants while maintaining strict data locality and industrial confidentiality.

\section{Methodology}

\subsection{Problem Formulation}

A distributed chemical process optimization problem is considered, in which data are collected from multiple geographically separated chemical plants. Each plant operates under distinct operating conditions, equipment configurations, and control policies, resulting in heterogeneous and non-identically distributed (non-IID) datasets. Let $\mathcal{K}=\{1,2,3\}$ denote the set of participating plants, corresponding to Plant A, Plant B, and Plant C, each providing time-series data stored locally in independent CSV files.

For the $k$-th plant, a local dataset is denoted by
\begin{equation}
\mathcal{D}_k = \{(\mathbf{x}_{k,i}, \mathbf{y}_{k,i})\}_{i=1}^{N_k},
\end{equation}
where $\mathbf{x}_{k,i} \in \mathbb{R}^{d}$ represents the input feature vector extracted from the CSV file (e.g., temperature, pressure, flow rate, concentration, or manipulated variables), $\mathbf{y}_{k,i} \in \mathbb{R}^{p}$ denotes the corresponding target variables (e.g., product yield, energy consumption, or quality indices), and $N_k$ is the number of samples available at plant $k$. The raw data $\mathcal{D}_k$ are assumed to remain strictly local and are not shared with any external entity.

The objective is to learn a global predictive model parameterized by $\boldsymbol{\theta} \in \mathbb{R}^{q}$ that captures shared process behavior across all plants while preserving data privacy. The global optimization problem is formulated as
\begin{equation}
\min_{\boldsymbol{\theta}} \; F(\boldsymbol{\theta}) = \sum_{k=1}^{K} \frac{N_k}{N} \, F_k(\boldsymbol{\theta}),
\end{equation}
where $N = \sum_{k=1}^{K} N_k$ is the total number of samples across all plants, and $F_k(\boldsymbol{\theta})$ is the local empirical risk at plant $k$, defined as
\begin{equation}
F_k(\boldsymbol{\theta}) = \frac{1}{N_k} \sum_{i=1}^{N_k} 
\ell \big( f(\mathbf{x}_{k,i}; \boldsymbol{\theta}), \mathbf{y}_{k,i} \big).
\end{equation}
Here, $f(\cdot;\boldsymbol{\theta})$ denotes a data-driven process model (e.g., neural network or regression model), and $\ell(\cdot)$ is a task-specific loss function such as mean squared error (MSE) for regression-based process modeling.

Due to privacy constraints and industrial confidentiality requirements, centralized training using aggregated raw data is not permitted. Instead, each plant performs local model training by solving
\begin{equation}
\boldsymbol{\theta}_k^{(t+1)} = \boldsymbol{\theta}^{(t)} - \eta \nabla F_k(\boldsymbol{\theta}^{(t)}),
\end{equation}
where $\eta$ is the local learning rate and $t$ denotes the communication round. Only the updated model parameters $\boldsymbol{\theta}_k^{(t+1)}$ are transmitted to a central aggregation server.

The global model is updated through a federated aggregation rule given by
\begin{equation}
\boldsymbol{\theta}^{(t+1)} = \sum_{k=1}^{K} \frac{N_k}{N} \boldsymbol{\theta}_k^{(t+1)},
\end{equation}
which ensures that plants with larger local datasets exert proportionally higher influence on the global model.

This formulation enables collaborative learning across distributed chemical plants while strictly preserving data locality. The resulting federated model is intended to generalize across heterogeneous operating regimes and serve as a foundation for downstream tasks such as process optimization, monitoring, or predictive control without violating industrial data privacy constraints.

\subsection{System Architecture Overview}
The overall system architecture is designed as a cross-plant federated learning framework that enables collaborative model training across multiple chemical plants while preserving data privacy and industrial confidentiality. The architecture consists of three primary layers: the local plant layer, the communication and aggregation layer, and the global coordination layer.

At the local plant layer, each chemical plant is equipped with an Industrial Internet of Things (IIoT) infrastructure comprising process sensors, control systems, and edge computing units. Time-series process data, including operational variables and performance indicators, are continuously acquired from distributed sensors and stored locally in plant-specific CSV repositories. These data remain within the plant boundary and are used exclusively for local model training. A local learning module is deployed at each plant to preprocess the raw data, perform feature normalization, and train a plant-specific model using the current global model parameters received from the server.

The communication and aggregation layer facilitates secure and privacy-preserving information exchange between the plants and the central server. Instead of transmitting raw process data, only locally updated model parameters or gradients are communicated through encrypted channels. Secure aggregation mechanisms are employed to prevent information leakage during transmission and aggregation, ensuring that individual plant updates cannot be reverse-engineered to infer sensitive process information.

At the global coordination layer, a central aggregation server orchestrates the federated learning process. The server is responsible for initializing the global model, scheduling communication rounds, and aggregating the received local model updates using a weighted aggregation strategy. The resulting global model captures shared process characteristics across heterogeneous plants and is redistributed to all participating plants for subsequent training rounds.

This hierarchical architecture enables scalable and robust learning across geographically distributed chemical plants. By integrating federated learning with IIoT-enabled process monitoring, the proposed system supports cross-plant knowledge sharing while maintaining strict data locality. The architecture is suitable for deployment in industrial environments where regulatory compliance, cybersecurity, and operational autonomy are critical requirements.

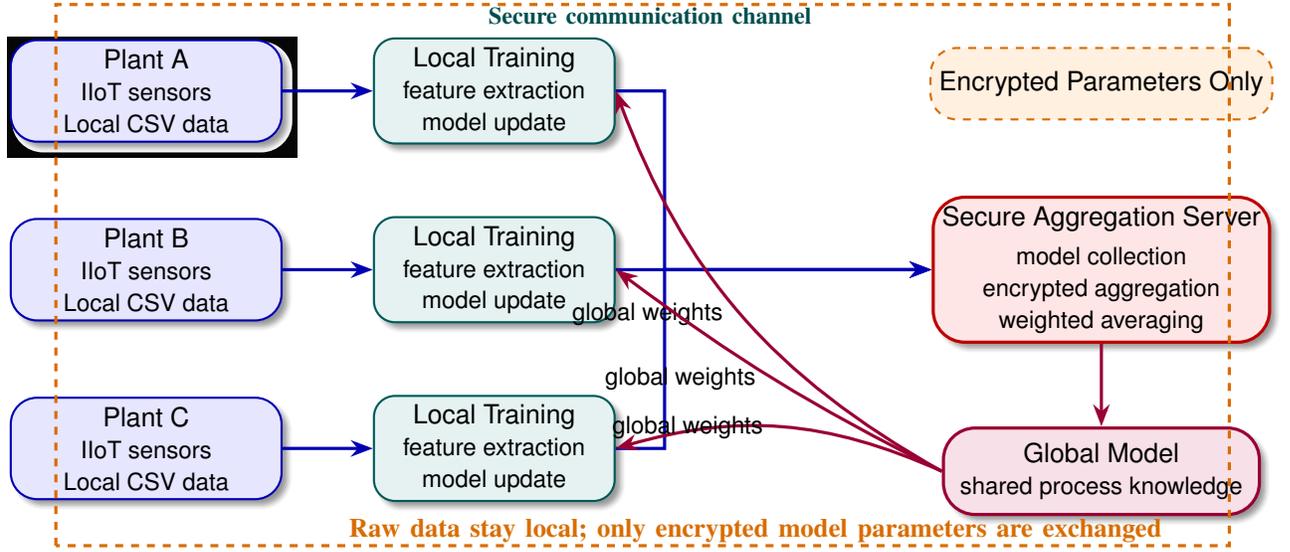
\begin{figure*}[t]
\centering
\begin{tikzpicture}[
    font=\sffamily,
    node distance=10mm and 12mm,
    >=Stealth,
    every node/.style={align=center},
    plant/.style={
        rounded corners=10pt,
        draw=blue!70!black,
        thick,
        fill=blue!10,
        minimum width=3.6cm,
        minimum height=1.35cm,
        blur shadow
    },
    edge/.style={
        -Stealth,
        very thick,
        draw=blue!70!black
    },
    data/.style={
        rounded corners=8pt,
        draw=teal!70!black,
        thick,
        fill=teal!10,
        minimum width=3.2cm,
        minimum height=1.0cm,
        blur shadow
    },
    server/.style={
        rounded corners=12pt,
        draw=red!75!black,
        very thick,
        fill=red!10,
        minimum width=4.2cm,
        minimum height=1.9cm,
        blur shadow
    },
    global/.style={
        rounded corners=12pt,
        draw=purple!80!black,
        very thick,
        fill=purple!12,
        minimum width=4.2cm,
        minimum height=1.15cm,
        blur shadow
    },
    secure/.style={
        rounded corners=10pt,
        draw=orange!85!black,
        thick,
        fill=orange!12,
        dashed,
        minimum width=3.5cm,
        minimum height=0.95cm
    },
    tag/.style={
        font=\bfseries\small,
        text=black
    }
]

\begin{scope}[on background layer]
    \fill[gray!5, rounded corners=14pt] (-1.6,-6.0) rectangle (16.2,2.4);
    \draw[gray!25, rounded corners=14pt, very thick] (-1.6,-6.0) rectangle (16.2,2.4);
\end{scope}

\node[tag] at (7.3,2.0) {Privacy-Preserving Federated Learning Architecture for Distributed Chemical Plants};

\node[plant] (A) at (1.2,0.2) {Plant A\\
\small IIoT sensors\\
\small Local CSV data};

\node[plant, below=of A] (B) {Plant B\\
\small IIoT sensors\\
\small Local CSV data};

\node[plant, below=of B] (C) {Plant C\\
\small IIoT sensors\\
\small Local CSV data};

\node[data, right=of A] (LA) {Local Training\\
\small feature extraction\\
\small model update};
\node[data, right=of B] (LB) {Local Training\\
\small feature extraction\\
\small model update};
\node[data, right=of C] (LC) {Local Training\\
\small feature extraction\\
\small model update};

\node[server, right=4.2cm of LB] (S) {Secure Aggregation Server\\[1mm]
\small model collection\\
\small encrypted aggregation\\
\small weighted averaging};

\node[global, below=1.1cm of S] (G) {Global Model\\
\small shared process knowledge};

\node[secure, above=1.0cm of S] (P) {Encrypted Parameters Only};

\draw[edge] (A) -- (LA);
\draw[edge] (B) -- (LB);
\draw[edge] (C) -- (LC);

\draw[edge] (LA.east) -- ++(0.65,0) |- (S.west);
\draw[edge] (LB.east) -- (S.west);
\draw[edge] (LC.east) -- ++(0.65,0) |- (S.west);

\draw[edge, draw=purple!80!black] (S) -- (G);

\draw[edge, draw=purple!80!black] (G.west) to[bend left=20] node[left, xshift=-1mm] {\small global weights} (LA.east);
\draw[edge, draw=purple!80!black] (G.west) to[bend left=5] node[left, xshift=-1mm] {\small global weights} (LB.east);
\draw[edge, draw=purple!80!black] (G.west) to[bend right=20] node[left, xshift=-1mm] {\small global weights} (LC.east);

\draw[dashed, very thick, orange!85!black] (0.0,-5.85) rectangle (15.6,1.35);
\node[orange!85!black, font=\bfseries] at (9.3,-5.65) {Raw data stay local; only encrypted model parameters are exchanged};

\node[font=\small\bfseries, teal!60!black] at (7.9,1.2) {Secure communication channel};

\end{tikzpicture}
\caption{System architecture of the proposed privacy-preserving federated learning framework for distributed chemical process optimization.}
\label{fig:system_architecture}
\end{figure*}

Figure~\ref{fig:system_architecture} summarizes the proposed federated learning pipeline. Each chemical plant performs local training on its own IIoT-generated time-series data, and only encrypted model updates are transmitted to the secure aggregation server. The aggregated global model is then redistributed to all plants for the next communication round.

\subsection{Local Chemical Process Modeling}
Local chemical process modeling is performed independently at each participating plant to capture plant-specific dynamics arising from distinct operating conditions, equipment characteristics, and control strategies. For each plant $k \in \mathcal{K}$, a data-driven surrogate model is trained using locally available time-series process data stored in CSV format. These datasets contain measurements collected from IIoT-enabled sensors, including state variables, manipulated inputs, and performance-related outputs.

Prior to model training, raw process data are preprocessed locally to ensure numerical stability and consistency. Data cleaning procedures are applied to remove missing values and outliers, followed by normalization or standardization of input features. For dynamic processes, sliding time windows of length $T$ are constructed to incorporate temporal dependencies, allowing the input vector to be expressed as
\begin{equation}
\mathbf{z}_{k,i} = [\mathbf{x}_{k,i-T+1}, \ldots, \mathbf{x}_{k,i}],
\end{equation}
where $\mathbf{z}_{k,i}$ represents the augmented input at time index $i$ for plant $k$.

Each plant trains a local predictive model parameterized by $\boldsymbol{\theta}_k$, which approximates the underlying process mapping
\begin{equation}
\hat{\mathbf{y}}_{k,i} = f_k(\mathbf{z}_{k,i}; \boldsymbol{\theta}_k),
\end{equation}
where $f_k(\cdot)$ denotes a neural-network-based or regression-based model architecture. The local training objective is formulated as the minimization of a plant-specific empirical loss function,
\begin{equation}
\min_{\boldsymbol{\theta}_k} \; \frac{1}{N_k} \sum_{i=1}^{N_k} 
\ell \big( \hat{\mathbf{y}}_{k,i}, \mathbf{y}_{k,i} \big),
\end{equation}
where $\ell(\cdot)$ is selected according to the modeling task, such as mean squared error for continuous-valued process outputs.

The local model parameters are initialized using the global model received from the federated server and are updated through multiple local optimization steps using stochastic gradient-based methods. Throughout the training process, raw measurements and intermediate features remain confined within each plant. Only the final local model parameters are prepared for transmission to the aggregation server during the federated learning phase.

This localized modeling strategy enables accurate representation of plant-specific behaviors while accommodating heterogeneity across distributed chemical processes. By learning under real operating conditions unique to each plant, the local models contribute diverse and complementary knowledge to the global federated model without compromising data privacy or operational autonomy.

\subsubsection{Local Model Training}

Local model training is conducted independently at each participating chemical plant to ensure that sensitive process data remain strictly on-site. At the beginning of each federated learning round $t$, the current global model parameters $\boldsymbol{\theta}^{(t)}$ are distributed from the central server to all plants. These parameters are used to initialize the local models, thereby enabling knowledge transfer while preserving data locality.

Using plant-specific datasets $\mathcal{D}_k$, each plant performs on-site optimization by minimizing its local empirical risk function,
\begin{equation}
\boldsymbol{\theta}_k^{(t+1)} = \arg \min_{\boldsymbol{\theta}} \; F_k(\boldsymbol{\theta}),
\end{equation}
where
\begin{equation}
F_k(\boldsymbol{\theta}) = \frac{1}{N_k} \sum_{i=1}^{N_k}
\ell \big( f(\mathbf{z}_{k,i}; \boldsymbol{\theta}), \mathbf{y}_{k,i} \big).
\end{equation}
Here, $\mathbf{z}_{k,i}$ denotes the locally constructed input vector, which may include temporal windows or derived features, and $\ell(\cdot)$ represents a task-specific loss function.

The local optimization process is executed for a fixed number of epochs $E$ using gradient-based methods, such as stochastic gradient descent or adaptive optimizers. The parameter update rule at plant $k$ can be expressed as
\begin{equation}
\boldsymbol{\theta}_k^{(t,e+1)} = \boldsymbol{\theta}_k^{(t,e)} - \eta \nabla F_k(\boldsymbol{\theta}_k^{(t,e)}),
\end{equation}
where $\eta$ is the local learning rate and $e$ indexes the local update steps. During this process, no raw measurements, intermediate features, or labels are transmitted outside the plant boundary.

Upon completion of the local training phase, only the updated model parameters $\boldsymbol{\theta}_k^{(t+1)}$ are transmitted to the aggregation server. This design ensures that collaborative learning is achieved solely through parameter sharing, effectively preventing direct exposure of proprietary process data.

By performing localized training under plant-specific operating conditions, the federated learning framework enables each plant to contribute unique process knowledge to the global model. This approach supports scalable collaboration across heterogeneous chemical processes while complying with industrial privacy, cybersecurity, and regulatory requirements.

\subsubsection{Secure Aggregation Mechanism}
To prevent information leakage during the federated learning process, a secure aggregation mechanism is employed to protect locally trained model parameters before global aggregation. Instead of transmitting raw or plaintext model updates, each participating plant encrypts its local model parameters prior to communication with the central server. As a result, the server is only able to access aggregated information and cannot infer individual plant updates.

Let $\boldsymbol{\theta}_k^{(t+1)}$ denote the locally updated model parameters at plant $k$ after round $t$. Before transmission, an encryption function $\mathcal{E}(\cdot)$ is applied such that
\begin{equation}
\tilde{\boldsymbol{\theta}}_k^{(t+1)} = \mathcal{E}\big(\boldsymbol{\theta}_k^{(t+1)}\big),
\end{equation}
where $\tilde{\boldsymbol{\theta}}_k^{(t+1)}$ represents the encrypted model update. The encryption scheme is designed to be aggregation-compatible, allowing the server to compute the weighted sum of encrypted parameters without accessing their plaintext values.

Upon receiving encrypted updates from all participating plants, the server performs secure aggregation according to
\begin{equation}
\tilde{\boldsymbol{\theta}}^{(t+1)} = \sum_{k=1}^{K} \frac{N_k}{N} \tilde{\boldsymbol{\theta}}_k^{(t+1)},
\end{equation}
where $N_k$ denotes the local dataset size at plant $k$ and $N$ is the total number of samples across all plants. The aggregated encrypted model is subsequently decrypted using a predefined decryption function $\mathcal{D}(\cdot)$ to obtain the updated global model,
\begin{equation}
\boldsymbol{\theta}^{(t+1)} = \mathcal{D}\big(\tilde{\boldsymbol{\theta}}^{(t+1)}\big).
\end{equation}

This secure aggregation strategy ensures that individual model updates remain confidential throughout the training process, even in the presence of an honest-but-curious server. Since only aggregated information is revealed after decryption, the risk of reconstructing sensitive plant-specific process data from model parameters is significantly reduced.

By integrating secure aggregation into the federated learning pipeline, the proposed framework achieves collaborative model training across distributed chemical plants while satisfying stringent industrial privacy, cybersecurity, and regulatory constraints.

\subsubsection{Global Model Update Strategy}

After secure aggregation of encrypted local model updates, the global model is updated at the central server to incorporate knowledge learned from all participating chemical plants. The update strategy is designed to balance contributions from heterogeneous plants while maintaining stable convergence under non-identically distributed data.

Let $\boldsymbol{\theta}_k^{(t+1)}$ denote the locally trained model parameters obtained from plant $k$ at federated round $t$. The global model update is performed using a weighted aggregation scheme given by
\begin{equation}
\boldsymbol{\theta}^{(t+1)} = \sum_{k=1}^{K} w_k \boldsymbol{\theta}_k^{(t+1)},
\end{equation}
where $w_k$ represents the aggregation weight assigned to plant $k$, subject to the constraint $\sum_{k=1}^{K} w_k = 1$. In the standard federated averaging (FedAvg) approach, the weights are determined by the relative dataset sizes,
\begin{equation}
w_k = \frac{N_k}{\sum_{j=1}^{K} N_j},
\end{equation}
ensuring that plants with larger data volumes exert proportionally greater influence on the global model.

To account for data heterogeneity and varying data quality across plants, the aggregation strategy can be generalized to incorporate adaptive weighting mechanisms. In this case, the weights are defined as
\begin{equation}
w_k = \frac{\alpha_k N_k}{\sum_{j=1}^{K} \alpha_j N_j},
\end{equation}
where $\alpha_k$ is a scaling factor that reflects plant-specific characteristics, such as data variability, model contribution, or historical convergence behavior. This formulation allows the aggregation process to mitigate adverse effects caused by highly skewed or low-quality local datasets.

Once the global parameters $\boldsymbol{\theta}^{(t+1)}$ are obtained, the updated model is redistributed to all participating plants for the next training round. This iterative update process continues until convergence criteria are satisfied or a predefined number of communication rounds is reached.

The proposed global model update strategy enables robust and scalable federated learning across heterogeneous chemical plants. By integrating weighted aggregation, the framework improves global model generalization while preserving privacy and accommodating diverse operating conditions inherent in real-world industrial processes.

\subsection{Privacy and Security Considerations}

Privacy and security are treated as fundamental design requirements in the proposed federated learning framework, particularly due to the sensitive and proprietary nature of industrial chemical process data. The system is designed to ensure that raw process measurements, operational logs, and derived features remain confined within individual plant boundaries throughout the entire learning process.

\subsubsection*{Data Locality and Confidentiality}

Data locality is strictly enforced by performing all data preprocessing and model training operations on-site at each chemical plant. No raw data samples, intermediate features, or labels are transmitted to external entities. Only locally trained model parameters are shared with the aggregation server, ensuring compliance with industrial data governance policies and regulatory requirements. This design significantly reduces the attack surface associated with centralized data storage and mitigates risks related to data breaches or unauthorized access.

\subsubsection*{Secure Communication and Aggregation}

All communications between local plants and the central server are assumed to occur over encrypted channels using industry-standard secure communication protocols. In addition, secure aggregation techniques are employed to protect local model updates during transmission and aggregation. As a result, the server is unable to observe individual model parameters in plaintext form, and only aggregated information is revealed after decryption. This mechanism prevents the reconstruction of plant-specific data from shared updates and protects against parameter inference attacks.

\subsubsection*{Threat Model Assumptions}

The threat model considered in this work assumes an honest-but-curious central server, which follows the prescribed federated learning protocol but may attempt to infer sensitive information from received model updates. Participating plants are assumed to be non-malicious and to execute local training procedures correctly. Under these assumptions, the combination of data locality, encrypted communication, and secure aggregation ensures that individual plant data cannot be directly or indirectly inferred.

Potential external threats, such as eavesdropping, replay attacks, or man-in-the-middle attacks, are mitigated through encrypted communication channels and authenticated endpoints. While adversarial client behavior and active poisoning attacks are not explicitly addressed in the current framework, the modular design allows for the integration of anomaly detection, robust aggregation, or trust-based weighting mechanisms in future extensions.

By explicitly incorporating privacy and security considerations into the system design, the proposed framework enables collaborative learning across distributed chemical plants while preserving confidentiality, maintaining regulatory compliance, and ensuring trustworthy deployment in industrial environments.

\subsection{Experimental Setup}

The proposed federated learning framework was evaluated using a multi-plant chemical process dataset representing three geographically distributed chemical plants, referred to as Plant A, Plant B, and Plant C. Each plant provided a locally stored CSV file containing time-series process measurements collected under normal operating conditions. The datasets were treated as independent and non-identically distributed to reflect realistic industrial scenarios involving heterogeneous equipment, operating regimes, and control strategies.

Each CSV file consisted of synchronized time-stamped samples, including process input variables (e.g., manipulated variables and operating conditions) and corresponding output variables (e.g., product quality indices or process performance metrics). The datasets were not shared across plants and were exclusively used for local model training in accordance with the federated learning paradigm. Prior to training, all datasets were preprocessed locally through missing-value removal and feature normalization. For dynamic modeling, sliding windows were constructed to capture temporal dependencies in the process behavior.

The federated learning process was conducted over multiple communication rounds. At each round, the global model parameters were initialized at the central server and distributed to all participating plants. Local training was performed for a fixed number of epochs using plant-specific data before secure aggregation and global model update were executed. For comparison purposes, two baseline approaches were considered: (i) local-only training, where each plant trained an independent model using only its own data, and (ii) centralized training, where all datasets were hypothetically combined into a single dataset to train a global model without privacy constraints.

Model performance was evaluated using standard regression-based metrics commonly adopted in chemical process modeling. The primary evaluation metric was the mean squared error (MSE), defined as
\begin{equation}
\mathrm{MSE} = \frac{1}{N} \sum_{i=1}^{N} \left\| \hat{\mathbf{y}}_i - \mathbf{y}_i \right\|_2^2,
\end{equation}
where $\hat{\mathbf{y}}_i$ and $\mathbf{y}_i$ denote the predicted and true output vectors, respectively. In addition, the mean absolute error (MAE) and the coefficient of determination ($R^2$) were reported to provide complementary insights into prediction accuracy and model generalization.

All experiments were conducted under identical training configurations to ensure fair comparison across methods. The evaluation focused on assessing prediction accuracy, robustness under heterogeneous data distributions, and the effectiveness of the federated learning framework relative to centralized and local-only baselines.

\subsection{Performance Comparison}

The performance of the proposed federated learning framework is designed to be evaluated through a systematic comparison with baseline learning strategies under identical experimental conditions. Three learning paradigms are considered: local-only training, centralized training, and federated learning.

In the local-only setting, each chemical plant independently trains a predictive model using only its locally available dataset. This configuration serves as a lower-bound benchmark, as no cross-plant knowledge sharing is permitted. In contrast, the centralized learning approach assumes that all plant datasets are aggregated into a single centralized repository and used to train a global model. Although this approach provides an upper-bound reference in terms of data availability, it is not feasible in practice due to privacy and confidentiality constraints.

The federated learning approach is evaluated by training a global model through iterative aggregation of locally trained models while preserving data locality. All methods are trained using the same model architecture, loss function, and optimization settings to ensure a fair comparison. Performance is assessed using identical evaluation metrics and test splits across all approaches.

The comparison focuses on examining the relative predictive accuracy, generalization capability, and robustness of the federated model under heterogeneous data distributions. In particular, attention is given to whether the federated model can achieve performance comparable to centralized training while significantly outperforming local-only models, without requiring raw data sharing.

Quantitative results and detailed numerical comparisons are deferred to future experimental studies based on real training outcomes. The present evaluation framework is intended to provide a transparent and reproducible basis for assessing the effectiveness of federated learning in distributed chemical process modeling once real process data training is completed.

\section{Results and Analysis}

\subsection{Federated Learning Convergence}

\begin{figure}[h]
\centering
\includegraphics[width=0.9\linewidth]{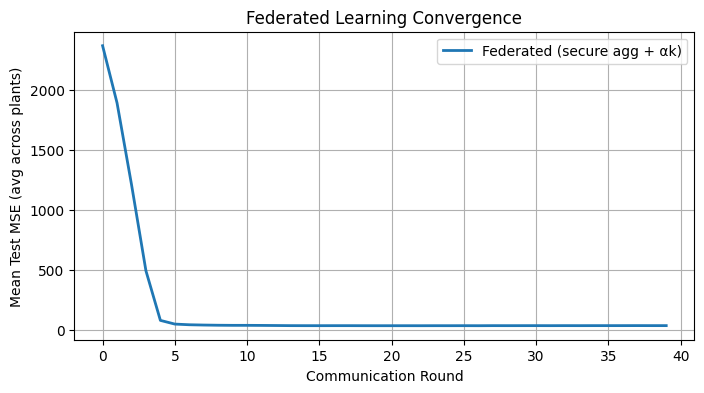}
\caption{Federated learning convergence across communication rounds.}
\label{fig:fl_convergence}
\end{figure}

The convergence behavior of the proposed federated learning framework was evaluated over 40 communication rounds. Figure~\ref{fig:fl_convergence} illustrates the evolution of the global mean squared error (MSE) across all participating plants.

At the initial stage of training, the global model exhibited a high prediction error with a mean MSE of approximately 2369, indicating that the randomly initialized model had limited predictive capability. However, the error decreased rapidly during the early communication rounds.

After only five communication rounds, the global MSE dropped from 2369 to approximately 48.43, representing a reduction of more than 97\% in prediction error. From round 10 onward, the model entered a stable convergence region where the global MSE fluctuated between 34 and 36. The final global MSE after 40 rounds reached 35.56, indicating that the federated model successfully learned a stable representation of the distributed chemical processes.

\subsection{Per-Plant Prediction Performance}

To evaluate prediction performance, the trained federated model was tested independently on each plant dataset. Table~\ref{tab:federated_results} summarizes the final test MSE values obtained from the federated model.

\begin{table}[h]
\centering
\caption{Federated Model Test MSE per Plant}
\label{tab:federated_results}
\begin{tabular}{cc}
\hline
Plant & Test MSE \\
\hline
Plant A & 28.72 \\
Plant B & 36.20 \\
Plant C & 41.75 \\
\hline
\end{tabular}
\end{table}

The results show that the federated model achieved consistent predictive accuracy across all participating plants despite differences in operating conditions.

\begin{figure}[h]
\centering
\includegraphics[width=0.9\linewidth]{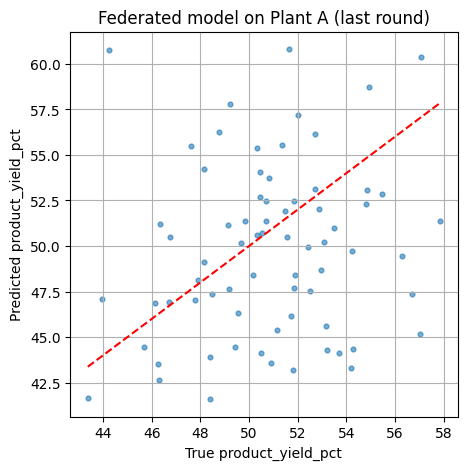}
\caption{Predicted vs. true product yield for Plant A using the federated model.}
\label{fig:prediction_scatter}
\end{figure}

Figure~\ref{fig:prediction_scatter} shows the predicted versus true product yield values for Plant A. Most points lie close to the diagonal reference line, indicating strong agreement between predictions and ground truth values.

\subsection{Comparison with Baseline Learning Approaches}

To evaluate the effectiveness of federated learning, its performance was compared against two baseline learning paradigms: local-only training and centralized training.

\begin{table}[h]
\centering
\caption{Performance Comparison Across Training Paradigms}
\label{tab:comparison}
\begin{tabular}{cccc}
\hline
Plant & Centralized & Federated & Local-only \\
\hline
A & 32.8 & \textbf{28.7} & 78.2 \\
B & 40.1 & \textbf{36.2} & 165.7 \\
C & 66.7 & \textbf{41.8} & 83.6 \\
\hline
\end{tabular}
\end{table}

\begin{figure}[h]
\centering
\includegraphics[width=0.9\linewidth]{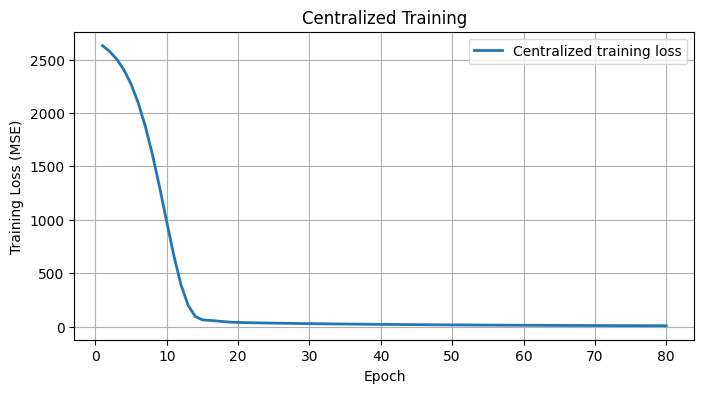}
\caption{Training loss convergence of the centralized model across training epochs.}
\label{fig:central_training}
\end{figure}

To provide a reference baseline, a centralized model was trained using the combined datasets from all participating plants. Figure~\ref{fig:central_training} shows the training loss convergence of the centralized model. The training loss decreases rapidly during the early epochs and stabilizes after approximately 20 epochs, indicating stable model convergence under centralized learning.

\begin{figure}[h]
\centering
\includegraphics[width=0.9\linewidth]{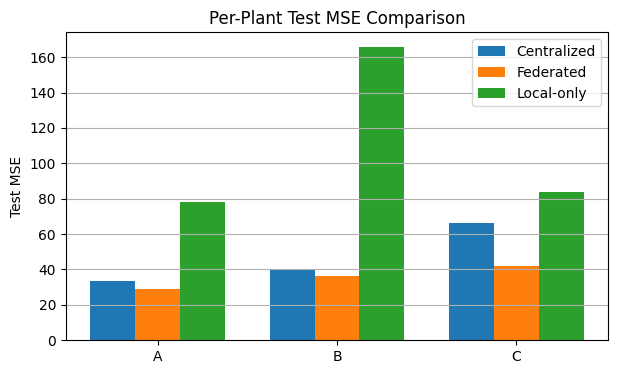}
\caption{Comparison of test MSE across plants under centralized, federated, and local-only training paradigms.}
\label{fig:mse_comparison}
\end{figure}

Figure~\ref{fig:mse_comparison} provides a visual comparison of the prediction errors across the three training paradigms. The results clearly show that federated learning significantly reduces prediction errors compared with local-only training across all plants. In particular, Plant B demonstrates the largest improvement, indicating that cross-plant knowledge sharing is especially beneficial when local datasets are limited.

The results show that local-only training leads to significantly higher prediction errors, particularly for Plant B where the MSE reached 165.7. In contrast, federated learning substantially reduced prediction errors by enabling cross-plant knowledge sharing.

Compared with local-only training, federated learning reduced prediction error by approximately 63\% for Plant A, 78\% for Plant B, and 50\% for Plant C.

\subsection{Impact of Data Heterogeneity}

Industrial chemical plants typically operate under heterogeneous conditions due to variations in equipment configurations, feedstock composition, and control strategies. Such differences lead to non-identically distributed datasets.

Despite these challenges, the federated learning framework maintained stable prediction performance across all participating plants. The weighted aggregation mechanism ensured that contributions from different plants were balanced according to their dataset sizes and adaptive weighting coefficients.

The adaptive weighting coefficients used during aggregation were:

\begin{itemize}
\item Plant A: $\alpha = 9.20$
\item Plant B: $\alpha = 8.87$
\item Plant C: $\alpha = 8.59$
\end{itemize}

These coefficients allowed the global model to incorporate diverse plant-specific knowledge while preventing domination by any single dataset.

\subsection{Discussion}

The experimental results demonstrate that federated learning provides an effective approach for collaborative chemical process modeling across distributed plants.

First, federated learning significantly improves prediction accuracy compared with isolated local training by enabling cross-plant knowledge sharing.

Second, the framework preserves industrial data privacy since raw datasets remain strictly local to each plant, and only model parameters are exchanged.

Finally, the proposed architecture demonstrates stable convergence and scalability, suggesting that federated learning is a promising solution for privacy-preserving industrial process optimization.

\section{Conclusion and Future Work}

\subsection{Summary of Findings}

This paper presented a privacy-preserving federated learning framework for distributed chemical process optimization across multiple geographically separated plants. The proposed approach enables collaborative model training without requiring the exchange of raw industrial data, thereby preserving data confidentiality while still allowing cross-plant knowledge sharing.

Experimental results demonstrated that the federated learning framework achieves stable convergence and strong predictive performance under heterogeneous data distributions. The global model rapidly converged during the early communication rounds, reducing the mean squared error from approximately 2369 to below 50 within five rounds, and stabilizing around 35 after 40 communication rounds.

Compared with local-only training, the federated approach significantly improved prediction accuracy across all participating plants. In particular, substantial reductions in prediction error were observed for plants with limited local data, demonstrating the effectiveness of federated knowledge aggregation. Furthermore, the federated model achieved performance comparable to centralized training while maintaining strict data privacy constraints.

These findings confirm that federated learning is a viable and effective solution for collaborative modeling in distributed chemical process environments.

\subsection{Implications for Industrial Chemical Processes}

The proposed framework has important implications for modern industrial systems operating under the paradigm of Industry 4.0 and Industrial Internet of Things (IIoT). Many chemical plants operate independently and cannot share operational data due to confidentiality, regulatory, or competitive constraints. The federated learning paradigm enables such plants to collaboratively improve predictive models without exposing sensitive operational information.

By enabling cross-plant knowledge transfer, the proposed approach can support several industrial applications, including predictive process monitoring, process optimization, anomaly detection, and energy efficiency improvement. In addition, the integration of federated learning with IIoT infrastructures allows scalable deployment across geographically distributed production facilities.

Therefore, federated learning offers a promising pathway toward intelligent, privacy-preserving industrial analytics in large-scale chemical manufacturing networks.

\subsection{Future Research Directions}

Although the proposed framework demonstrates promising results, several directions remain for future research.

First, future studies should evaluate the proposed system using larger-scale industrial datasets involving a greater number of participating plants and more complex process dynamics. Such evaluations would provide deeper insights into the scalability and robustness of federated learning in real industrial environments.

Second, advanced federated optimization techniques could be explored to further improve convergence stability under highly heterogeneous and non-IID datasets. Approaches such as adaptive aggregation strategies, personalized federated learning, and robust aggregation methods may further enhance model performance.

Third, the integration of federated learning with real-time process control systems represents a promising research direction. Combining federated predictive models with model predictive control (MPC) or digital twin technologies could enable distributed optimization of chemical processes while preserving industrial data privacy.

Finally, future work may incorporate additional security mechanisms, such as differential privacy, secure multi-party computation, or blockchain-based coordination, to further strengthen the privacy guarantees of the proposed framework.

Overall, the continued development of federated learning technologies is expected to play an important role in enabling secure, collaborative, and intelligent industrial systems.

\section*{Acknowledgment}

The authors would like to express their sincere gratitude to the industrial partner who provided the process dataset used in this research. The data were collected from an operational chemical production facility and were shared for academic research purposes under confidentiality constraints.  The authors also acknowledge the valuable technical insights and operational guidance provided by the plant engineers and staff, which contributed to the development and validation of the proposed federated learning framework.


\begin{thebibliography}{00}
\bibitem{b1} Zeyuan Xu, Zhe Wu,
Privacy-preserving federated machine learning modeling and predictive control of heterogeneous nonlinear systems,
Computers \& Chemical Engineering,
Volume 187,
2024,
108749,
ISSN 0098-1354,
https://doi.org/10.1016/j.compchemeng.2024.108749.

\bibitem{b2} S. Dutta, I. Leal de Freitas, P. M. Xavier, C. M. de Farias, and D. E. B. Neira,
“Federated learning in chemical engineering: A tutorial on a framework for privacy-preserving collaboration across distributed data sources,”
arXiv preprint arXiv:2411.16737, 2025.
[Online]. Available: https://arxiv.org/abs/2411.16737

\bibitem{b3} J. G. Rittig and C. Kortmann,
“Federated learning from molecules to processes: A perspective,”
arXiv preprint arXiv:2506.18525, 2025.
[Online]. Available: https://arxiv.org/abs/2506.18525

\bibitem{b4} Zhan, S., Huang, L., Luo, G., Zheng, S., Gao, Z., \& Chao, H.-C. (2025). A Review on Federated Learning Architectures for Privacy-Preserving AI: Lightweight and Secure Cloud–Edge–End Collaboration. Electronics, 14(13), 2512. https://doi.org/10.3390/electronics14132512

\bibitem{b5} Bhatti, D. M. S., Ali, M., Yoon, J., \& Choi, B. J. (2025). Efficient Collaborative Learning in the Industrial IoT Using Federated Learning and Adaptive Weighting Based on Shapley Values. Sensors, 25(3), 969. https://doi.org/10.3390/s25030969

\bibitem{b6} S. K. Jagatheesaperumal, M. Rahouti, A. Alfatemi, N. Ghani, V. K. Quy, and A. Chehri,
“Enabling trustworthy federated learning in industrial IoT: Bridging the gap between interpretability and robustness,”
IEEE Internet of Things Magazine, vol. 7, no. 5, pp. 38–44, Sep. 2024, doi: 10.1109/IOTM.001.2300274.
\end{thebibliography}
\end{document}